%% file: acl.tex
\pdfoutput=1

\documentclass[11pt]{article}

\usepackage{acl}

\usepackage{times}
\usepackage{latexsym}

\usepackage[T1]{fontenc}

\usepackage[utf8]{inputenc}

\usepackage{microtype}

\usepackage{inconsolata}

\usepackage{amsmath,amsfonts}
\usepackage{booktabs}
\usepackage{graphicx}
\usepackage{caption}
\usepackage{subcaption}
\usepackage[ruled,vlined]{algorithm2e}
\usepackage{algorithmicx}
\usepackage{algpseudocode}
\usepackage{makecell,tabularx}
\usepackage{soul}

\usepackage{tikz}
\usepackage{diagbox}
\usepackage{multirow}
\usepackage{float}
\usepackage{tablefootnote}
\usepackage{bbm}
\usepackage{mathtools}
\usepackage{enumitem}

\definecolor{lightblue}{RGB}{173,216,230}

\mathtoolsset{showonlyrefs,showmanualtags}

\newcommand{\comment}[1]{}

\usepackage{pgfplots}
\pgfplotsset{compat=1.18}
\usetikzlibrary{patterns}
\usepackage{tikz}
\usetikzlibrary{positioning}

\definecolor{bananamania}{rgb}{0.98, 0.91, 0.71}
\definecolor{babyblue}{rgb}{0.54, 0.81, 0.94}
\definecolor{celadon}{rgb}{0.67, 0.88, 0.69}
\definecolor{coral}{rgb}{1.0, 0.5, 0.31}
\definecolor{coolgrey}{rgb}{0.55, 0.57, 0.67}
\definecolor{ashgrey}{rgb}{0.7, 0.75, 0.71}
\definecolor{bittersweet}{rgb}{1.0, 0.44, 0.37}

\renewenvironment{quote}{%
  \list{}{%
    \leftmargin0.1cm   %
    \rightmargin\leftmargin
  }
  \item\relax
}
{\endlist}

\newenvironment{tightquote}
{\vspace{-0.4\baselineskip}\quote}
{\endquote\vspace{-0.4\baselineskip}}

\title{Can Large Language Model Summarizers Adapt to\\ Diverse Scientific Communication Goals?}

\author{Marcio Fonseca $\qquad$ Shay B. Cohen \\
Institute for Language, Cognition and Computation \\
School of Informatics, University of Edinburgh \\
10 Crichton Street, Edinburgh, EH8 9AB \\
\medskip
\texttt{m.fonseca@ed.ac.uk},
\texttt{scohen@inf.ed.ac.uk}}

\begin{document}
\maketitle
\begin{abstract}

In this work, we investigate the controllability of large language models (LLMs) on scientific summarization tasks. We identify key stylistic and content coverage factors that characterize different types of summaries such as paper reviews, abstracts, and lay summaries. By controlling stylistic features, we find that non-fine-tuned LLMs outperform humans in the MuP review generation task, both in terms of similarity to reference summaries and human preferences. Also, we show that we can improve the controllability of LLMs with keyword-based classifier-free guidance (CFG) while achieving lexical overlap comparable to strong fine-tuned baselines on arXiv and PubMed. However, our results also indicate that LLMs cannot consistently generate long summaries with more than 8 sentences. Furthermore, these models exhibit limited capacity to produce highly abstractive lay summaries.  Although LLMs demonstrate strong generic summarization competency, sophisticated content control without costly fine-tuning remains an open problem for domain-specific applications.\footnote{Code is available at \url{https://github.com/thefonseca/scisum}}

\end{abstract}

\section{Introduction}

\begin{figure}[ht!]
  \centering
  \scalebox{0.88}{
  \begin{tabularx}{0.54\textwidth}{X}
    \includegraphics[trim={1.6cm 6.05cm 12.9cm 3cm},clip,width=0.52\textwidth,height=8cm]{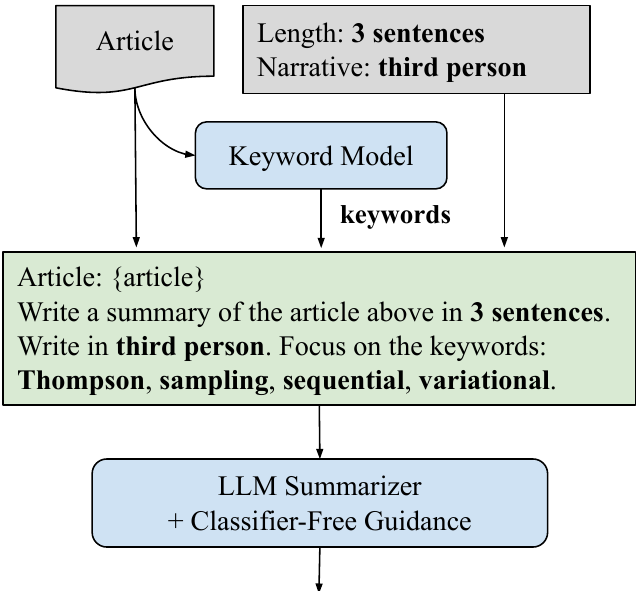}\\
    \midrule
    \textbf{SUMMARY}: This work develops an analog to \textbf{\hl{Thompson}} \textbf{\hl{sampling}} by upper-bounding the expected regret in \textbf{\hl{sequential}} decision-making problems. The two terms in the upper bound loosely resemble the evidence lower bound of \textbf{\hl{variational}} inference: the first term encourages selecting arms with high expected reward; the second term depends on the inverse of the rate function and penalizes heavy tails and encourages exploration. The resulting \textbf{\hl{variational}} \textbf{\hl{Thompson}} \textbf{\hl{sampling}} algorithm is evaluated on a random game and a constrained bandit problem.\\
    \bottomrule
  \end{tabularx}
  }

\caption{An overview of our controllability experiments. We expose LLM summarizers to prompts conveying communicative intentions related to \emph{conciseness}, \emph{narrative perspective}, and \emph{keywords} inferred by a \emph{keyword model}. Then, we measure how generated summaries adhere to those intentional targets.\vspace{-0.35cm}}\label{fig:overview}
\end{figure}

Recent work on the evaluation of large language models (LLMs) has shown unprecedented performance on diverse language generation tasks, even in zero-shot settings \citep{clark_all_2021}. Specifically in text summarization, \citet{goyal2022news} found that human evaluators prefer summaries produced by GPT-3 over strong supervised baselines. In a similar experiment testing ten large language models, \citet{zhang_benchmarking_2023} concluded that instruction-tuned models performed on par with human freelance writers. Additionally, their results suggest that the evaluation of news summarization is hindered by the low quality of reference summaries. Hence, the conventional summarization paradigm based on fine-tuning and evaluating on reference datasets is questionable in the face of increasingly competent language models.

However, most of those studies highlighting the success of LLM summarizers are based on news articles and the validity of those claims for long technical documents deserves further investigation. In this work, we aim to fill this gap by posing the following research questions: 1) How do LLM summaries compare to human-generated paper summaries? 2) To what extent LLM summarizers can be controlled to fulfill different goals of scientific communication? In both cases, we focus on the zero-shot setting, since the cost to fine-tune such large models is often prohibitive.

To address the first question, we design an experiment to compare human and machine summarizers, in which human summaries are judged in terms of their similarity with respect to another set of human-written reference summaries. By evaluating the lexical overlap with reference summaries from the multi-perspective scientific summarization dataset - MuP \citep{cohan-etal-2022-overview-first}, we find that human reviewers achieve lower ROUGE scores \citep{lin-2004-rouge} compared to LLMs. The high quality of LLM summaries is confirmed by a human judgment experiment, where machine-generated summaries are preferred in 83\% of the instances.

The strong preference for LLM summaries hints that the variability in communicative intentions \citep{giulianelli2023comes, andreas2022language} outweighs the usual quality criteria such as coherence, fluency, and relevance \citep{fabbri2021summeval}. In fact, an inspection of human summaries for the same article reveals arbitrary decisions related to conciseness and coverage of scientific aspects. We argue that those subjective decisions affect the perceived quality of summaries and that evaluation protocols should consider the \emph{adaptability} of summarizers to diverse contexts \citep{jones1998automatic}.

Motivated by the experiment on the MuP dataset and previous literature on controllable summarization \citep{he-etal-2022-ctrlsum}, we elect three intentional aspects: conciseness, narrative perspective (first or third person), and keyword coverage. For each of those aspects, we define two elements: \emph{intention prompts}, and \emph{intention control metrics}. Intentional prompts are designed to specify the task and to elicit a given intentional behavior from LLMs (e.g., \emph{summarize this article in \textbf{5 sentences}}). Finally, the control metrics assess the intentional alignment of prompts and the generated summaries.

To investigate the second research question, we evaluate our intentional prompts on the tasks of abstract generation \citep{Cohan_2018_discourse_aware} and lay summarization \citep{goldsack-etal-2022-making}, using \textsc{GPT-3.5} and \textsc{Llama-2} \citep{touvron2023llama}. We find that both models can follow conciseness and narrative perspective intentions accurately to generate short abstracts for arXiv and PubMed. Furthermore, by using keyword-based intentional prompts and classifier-free guidance \citep{Sanchez2023StayOT}, we can direct to the lexical content of summaries leading to ROUGE scores comparable to strong supervised baselines.

However, the eLife lay summarization benchmark presents a harder challenge for LLMs. Besides being highly abstractive, these lay summaries are much longer (18 sentences on average), and have a particular concept distribution emphasizing research background. Our controllability results show that intentional prompts achieve limited success in replicating these characteristics.

Overall, our experiments spanning four benchmarks indicate that LLMs are effective summarizers for long scientific documents, both in terms of lexical alignment and human preferences. However, when summarization tasks deviate from the training distribution (e.g., lay summarization), LLMs cannot consistently match the features of human summaries. Thus, LLMs do not usher the ``death'' of summarization \citep{pu2023summarization} but the transition from reference-based evaluation to more nuanced, domain-specific evaluation protocols, as envisioned by \citet{jones1998automatic} 25 years ago.

\section{Guiding Summarizer Intentions}
\label{sec:guidance}
As uncertainty in intentionality is intrinsic in language generation \citep{giulianelli2023comes}, we argue that summarization performance relates to the capacity of a system to adapt its behavior given not only the source documents but also the target communicative intentions.  We investigate language model intent adaptability by changing the summarization context via prompting, and measuring how it affects the perceived intention in the summaries (see Figure \ref{fig:overview} for an example). Let $P_{\theta}$ be a language model parameterized by $\theta$. Then, we define summarization as a conditional sequence generation: 
\begin{align}\label{eq:summary_generation}
S \sim P_{\theta}(\cdot \mid D, p_{I_1},\ldots,p_{I_N}),
\end{align}
where $D$ is an input article and $p_{I_1},\ldots,p_{I_N}$ are prompts inducing the intentions $I_1,\ldots,I_N$. In this work, we consider three types of intentions that are important to adapt scientific summaries to different goals, namely \emph{conciseness}, \emph{narrative perspective}, and \emph{keyword coverage}:

\begin{itemize}
    \item \textbf{Conciseness}: as one of the most important intentional factors in summarization, conciseness defines the compression rate between the source document and the summary. We are interested in the ability of LLMs to follow specific conciseness instructions ranging from short abstracts (6-8 sentences) to longer lay summaries (14 or more sentences).
    \item \textbf{Narrative Perspective}: depending on the perspective of the summarizer, the summary is written using first or third-person narrative. In our experiments, we consider that paper abstracts from arXiv and PubMed use first-person narrative, and other summaries such as paper reviews from MuP \citep{cohan-etal-2022-overview-first} and lay summaries from eLife \citep{goldsack-etal-2022-making} use third person narrative.
    \item \textbf{Keyword Coverage}: by guiding the coverage of keywords, we can indirectly manipulate the level of abstractiveness of a summary and favor simplified language, which is particularly relevant for lay summarization. Furthermore, when reference summaries are available, keyword guidance provides a mechanism to guide LLM generation using relatively small \emph{keyword models} (Section \ref{sec:keyword_model}) that demand fewer resources to fine-tune.
\end{itemize}

Since we specify intention guidance via prompts, we have no control over how the language models trade off their unconditional summarization behavior and the intention instructions. To address this issue, we modify the decoding process in Eq. \ref{eq:summary_generation} to include classifier-free guidance (CFG) weighting \citep{Ho2022ClassifierFreeDG,Sanchez2023StayOT}:
\begin{align}\label{eq:cfg}
\hat{P}_{\theta}(S | D, p_{I}, p_{\epsilon}) \propto \frac{P_{\theta}(S | D, p_{I})^\gamma}{P_{\theta}(S | D, p_{\epsilon})^{\gamma-1}},
\end{align}
where $\gamma \geq 1$ is the guidance strength and $p_\epsilon$ is a ``non-intentional'' summarization prompt:

\noindent\fbox{%
    \parbox{0.975\columnwidth}{%
\begingroup
\addtolength\leftmargin{-0.1in}
\begin{tightquote}
\texttt{Write a summary of the article above.}
\end{tightquote}
\endgroup
}}
When $\gamma > 1$, the next-token probabilities are changed so that the generated summary is closer to the target intentions than its default summarization behavior elicited by the prompt $p_\epsilon$.

To measure the adherence of summaries to intention prompts, we define reference-free \emph{intention control} metrics $k_I(y_I, S)$ that gauge the intentional alignment of $S$ with respect to a target value $y_I$. In the next sections, we detail the intention prompts $p_I$ and control metrics $k_I$ used in our experiments.

\subsection{Intention Prompt Templates}
\label{sec:prompt_design}
The next component of our methods refers to the way we prompt language models to induce the intentions described above. The general prompt template is defined as follows:

\noindent\fbox{%
    \parbox{0.975\columnwidth}{%
\begingroup
\addtolength\leftmargin{-0.1in}
\begin{tightquote}
\texttt{Article: \{text\}}\\
$\{p_{I_1} \ldots p_{I_N}\}$
\end{tightquote}
\endgroup
}}
where the placeholder \texttt{\{text\}} denotes the input article and $\{p_{I_1} \ldots p_{I_N}\}$ refers to the concatenation of intention prompts, namely conciseness, narrative perspective, and keyword coverage.

\subsubsection{Style Intention Prompts}
\label{sec:style_prompts}

\paragraph{Conciseness prompt}
We define an intention prompt for a target number of sentences $y_{\text{conciseness}}$:

\noindent\fbox{%
    \parbox{0.975\columnwidth}{%
\begingroup
\addtolength\leftmargin{-0.1in}

\begin{tightquote}
\texttt{Write a summary of the article above in $\{y_{\text{conciseness}}\}$ sentences.}
\end{tightquote}
\endgroup
}}
We choose to specify the target in sentences as it was found to be effective in previous work \citep{goyal2022news} and in our own experiments.

\paragraph{Narrative perspective prompt}
For paper abstract generation tasks, we use the following prompt to instruct the model to write in the usual first-person plural perspective:

\noindent\fbox{%
    \parbox{0.975\columnwidth}{%
\begingroup
\addtolength\leftmargin{-0.1in}

\begin{tightquote}
\texttt{Write in first person ``we'' when applicable.}
\end{tightquote}
\endgroup
}}
For other tasks, we do not have to prompt the model to use third person voice, as we observe this is the default behavior of the language models we evaluate (see Tables \ref{tab:abstract_generation} and \ref{tab:lay_summarization}).

\subsubsection{Keyword Coverage Prompt}
\label{sec:keyword_model}
The last prompt type instructs the model to focus on a collection of keywords to generate the summary:
\noindent\fbox{%
    \parbox{0.975\columnwidth}{%
\begingroup
\addtolength\leftmargin{-0.1in}

\begin{tightquote}
\texttt{Focus on the following keywords: $\{y_{\text{keywords}}\}$.}
\end{tightquote}
\endgroup
}}
where the $\{y_{\text{keywords}}\}$ placeholder indicates a comma-separated list of terms, which are provided by a \emph{keyword model}.

\paragraph{Keyword model}
In our experiments, we use pre-trained encoder-decoder summarizers as keyword generators. Specifically, we use \textsc{FactorSum} \citep{fonseca2022factorizing} and \textsc{Bart} \citep{lewis2019bart} for abstract generation and lay summarization respectively. Using these models, we generate summaries for each document in the evaluation dataset. Then, we extract part-of-speech tags for tokens in summaries using the NLTK library \citep{Bird_Natural_Language_Processing_2009} and keep as keywords only nouns, verbs, adjectives, foreign words, and symbols. The full list of keyword POS tags is presented in Appendix \ref{sec:keyword_model_details}. 

\subsection{Intention Control Metrics}
\label{sec:intention_metrics}
In this section, we identify the controllability metrics $k_I$ for stylistic features and keyword coverage. Although we present summary-level definitions, in our experiments we report dataset-level metrics, that is, their average over all evaluation samples.

\paragraph{Conciseness}
 In this work, we measure the conciseness of a summary $S$ by counting the number of sentences $|S|$. Then, the conciseness controllability $k_I$ for a summary $S$ and a target number of sentences $y_I$ is defined by:
\begin{align}
k_{\text{conciseness}}(y_I, S) = \texttt{abs}(y_I - |S|),
\end{align}
where \texttt{abs} is the absolute difference function.

\paragraph{Narrative perspective}
We define the narrative perspective metric based on a text classifier $f_{\text{narrative}}(s)$ that maps each summary sentence $s \in S$ to the label $\hat{y}_I \in \{\texttt{first}, \texttt{third}\}$. Then, we defined the summary-level perspective controllability as the percentage of sentences written in the target perspective $y_I$:
\begin{align}
k_{\text{narrative}}(y_I, S) = \frac{100}{|S|}\sum_{s_i \in S}\mathbbm{1}[f_I(s_i)=y_I],
\end{align}
where $|S|$ is the number of sentences in $S$. For the classifier $f_{\text{narrative}}(s)$, we use a simple heuristic where the sentence is classified as first-person perspective if 1) it starts with a first-person pronoun $\texttt{FPP} \in \{\texttt{``we''}, \texttt{``our''}\}$ or 2) if it contains the pattern \texttt{", FPP"}, i.e., a comma followed by a \texttt{FPP}.

\paragraph{Keyword Coverage}
We also measure the coverage of a target set of keywords $y_I$ in summaries using the ROUGE-1 recall metric \citep{lin-2004-rouge}:
\begin{align}
k_{\text{keywords}}(y_I, S) = \texttt{ROUGE-1}_{\text{recall}}(y_I, S).
\end{align}

\section{Experiments}
In this section, we present the experimental settings spanning three styles of scientific summaries: abstracts, reviews, and lay summaries. 

\subsection{Datasets}

\paragraph{Multi-perspective scientific summarization}
The MuP dataset \citep{cohan-etal-2022-overview-first} is a corpus of summaries extracted from publicly available scientific peer reviews\footnote{\url{https://openreview.net/}} capturing multiple summarization perspectives for a single document. In our experiments, we use 1,060 review summaries (and their corresponding papers) from the validation set, covering topics primarily from the AI, Machine Learning, and Natural Language Processing fields.

\paragraph{Abstract generation}
We use arXiv and PubMed \citep{Cohan_2018_discourse_aware}, which are two large-scale benchmarks for abstract generation. In our experiments, we report results for 1,000 random samples from each dataset test set.

\paragraph{Lay summarization}
We also evaluate LLMs on the task of lay summarization using the eLife dataset \citep{goldsack-etal-2022-making}. In contrast to reviews and abstracts, these summaries are much longer (around 18 sentences on average), more abstractive, and exhibit a strong bias towards background content, which makes them easier to parse by non-specialists. We report results on the 241 samples from the eLife test set.

\subsection{Models}

\paragraph{Supervised baselines}
For the abstract generation task, we use \textsc{BigBird} \citep{zaheer2020big}, a transformer-based model that uses a sparse attention mechanism to handle long input sequences. Also, we include \textsc{FactorSum} \citep{fonseca2022factorizing}, a model that breaks the summarization task into shorter subtasks, achieving strong performance on the arXiv and PubMed datasets. For lay summarization, we train our own version of the \textsc{BART} model \citep{lewis2019bart} on the eLife dataset (refer to Appendix \ref{sec:generation_details} for training details).

\paragraph{\textsc{Llama-2}}
A collection of pre-trained and fine-tuned large language models (LLMs) ranging in scale from 7 billion to 70 billion parameters \citep{touvron2023llama}. The intention prompts described in Section \ref{sec:prompt_design} are wrapped in a model-specific instruction-tuning prompt (see Appendix \ref{sec:generation_details} for details). Unless otherwise stated, we report results for the \emph{chat} variant with 7B parameters (16-bit floating point), with 4,096 maximum context tokens using nucleus sampling \citep{holtzman2019curious} with temperature $0.8$ and $p = 0.95$.

\paragraph{\textsc{GPT-3.5}}
A proprietary model based on \textsc{InstructGPT} \citep{ouyang2022training}. We use the model version \texttt{gpt-3.5-turbo-0301} via the chat completion API endpoint\footnote{\url{https://platform.openai.com/docs/api-reference/chat}}. Although we do not have access to the implementation details behind commercial APIs, they were extensively studied in recent work \citep{zhang_benchmarking_2023, goyal2022news}. This model version is trained on data up to September 2021, and we generate summaries in December 2023. 

\subsection{Evaluation Metrics}
In addition to the intention control metrics defined in Section \ref{sec:intention_metrics}, we report other relevant metrics for summarization, namely ROUGE \citep{lin-2004-rouge} and summary abstractiveness (percentage of n-grams in summary that are absent in the source document). For the lay summarization task, we also report the Flesch-Kincaid Grade Level (FKGL) readability score \citep{kincaid1975derivation}.

\section{Results and Discussion}
In this section, we discuss our experimental results comparing human and LLM-generated summaries (Section \ref{sec:human_vs_llms}), and the controllability experiments on the abstract generation and lay summarization tasks (Section \ref{sec:style_intentions}).

\subsection{Comparing Human and LLM Summaries: Reviewer Perspectives}
\label{sec:human_vs_llms}
We leverage the MuP dataset \citep{cohan-etal-2022-overview-first} for a controlled comparison between human and machine summaries. First, we select documents from the validation set with more than one human-written summary and randomly choose one of the summaries to serve as a reference and a second human-written summary as a fictitious system summary. This setting puts humans in a similar condition as usual reference-based summarization benchmarks, that is, humans are evaluated on their ability to \emph{guess} reference summaries. 

As machine summarizers, we evaluate \textsc{Llama-2} and \textsc{GPT-3.5}. To minimize the confounding factors related to summary length, we employ conciseness prompts (Section \ref{sec:style_prompts}) to instruct the LLMs to generate summaries with a number of sentences such that the number of generated tokens approximates the human summaries\footnote{We take the number of tokens in each human summary and divide by the average number of tokens per sentence generated by  \textsc{Llama-2} and \textsc{GPT-3.5}, resulting in a different conciseness target for each document.}. In Table \ref{tab:mup_eval}, we report metrics comparing LLM and human summaries for 1,060 samples from the MuP validation set.

In addition to automatic evaluation, we perform human evaluation based on 30 random samples of human and LLM-generated summaries. Similarly to \citet{goyal2022news}, we ask evaluators to perform blind A/B judgments based on the paper abstract and a triplet consisting of one human and two LLM summaries. Evaluators are tasked to elect from the triplet which summary (or summaries) they judge is the best/worst alternative to the paper abstract. For each answer, the evaluators need to provide a short justification. Each summary triplet is annotated by two graduate students specializing in NLP. We provide detailed task instructions in Appendix \ref{sec:human_evaluation}.

\paragraph{ROUGE penalizes abstractiveness in human summaries}
The results in Table \ref{tab:mup_eval} show that LLMs achieve comparable albeit higher ROUGE scores compared to human reviewers, which by standard reference-based protocols, would be an indicator that LLM summaries have ``superior quality.'' However, we observe quantitative differences between human and machine reviews across other metrics. Notably, human summaries are significantly more abstractive as measured by the percentage of novel bigrams in the summaries. Given the more creative nature of human reviews, it is expected that they offer more possibilities for lexical disagreement than a conservative summary that prioritizes content from the source document.

\begin{table}
  \centering
  \setlength\tabcolsep{1.2pt}
  \begin{tabular}{l|cccc}
    \toprule
    \textbf{Metric} & \textbf{ Ref } & \textbf{Human} & \textsc{\textbf{Llama}} & \textsc{\textbf{GPT-3.5}} \\
    \toprule
    ROUGE-1 & 100 & 35.55 & \textbf{37.35} & 36.39 \\
    ROUGE-2 & 100 & 8.12 & \textbf{9.20} & 8.51 \\
    ROUGE-L & 100 & 20.55 &  \textbf{24.47} & 24.02 \\
    \midrule
    Avg. Tokens & 115 & 113 & 112 & 113 \\
    Third Person & 99.8 & 99.6 & 99.3 & 100.0 \\
    Novel bigram\,\, & 66.4 & \textbf{67.0} & 39.9 & 34.1 \\
    \bottomrule
  \end{tabular}
  \caption{Metrics comparing human-written, \textsc{Llama-2-7B}, and \textsc{GPT-3.5} summaries to reference reviews (Ref) from the MuP validation set. We report average tokens per summary and the percentage of sentences using \emph{third person} perspective. \emph{Novel bigram} measures the percentage of novel bigrams in summaries.
  }\label{tab:mup_eval}
\end{table}

\paragraph{Humans strongly prefer LLM summaries}
The results in Table \ref{tab:mup_human_eval} reveal the marked preference of evaluators towards LLM summaries, which account for 83\% of votes for best summary. The preference for the worst summary is more balanced, with humans and \textsc{GPT-3.5} taking the majority of votes. Also, we note that Cohen's $\kappa$ scores \citep{artstein2008inter} are low, which indicates the high subjectivity of the task. 

To understand the factors underlying human preferences, we categorize the evaluators' comments into quality criteria, as shown in Table \ref{tab:mup_human_eval_reasons}. For both best and worst summary choices, the dominant reason is \emph{coverage} of research aspects such as background or experimental results. Interestingly, \emph{factuality} only appears as a third factor for choosing a bad summary, which is an indicator of the high quality of LLM summaries. However, we note that factuality issues might be underrepresented in our results, as the annotators do not use the original article in their judgments.

\begin{table}
  \centering
  \setlength\tabcolsep{2.1pt}
  \begin{tabular}{l|cccc}
    \toprule
    \textbf{Vote} & \textbf{Human} & \textbf{\textsc{Llama}} & \textbf{\textsc{GPT-3.5}} & \textbf{Cohen's $\kappa$} \\
    \toprule
    Best $\uparrow$ & 16.9 & 40.0 & \textbf{43.1} & 0.06 \\
    Worst $\downarrow$ & 36.1 & \textbf{27.9} & 36.1 & 0.17 \\
    \bottomrule
  \end{tabular}
  \caption{Human preferences (percentage) for best/worst MuP summaries. To account for multiple summary choices, we use the weighted Cohen's $\kappa$ agreement statistic \citep{artstein2008inter}.
  }\label{tab:mup_human_eval}
\end{table}

\begin{table}
  \centering
  \begin{tabular}{lclc}
    \toprule
    \textbf{Best} & \% & \textbf{Worst} & \% \\
    \toprule
    \textbf{Coverage} & \textbf{64.1} & \textbf{Coverage} & \textbf{42.9} \\
    Coherence & 12.8 & Relevance & 11.1 \\
    Fluency & 6.4 & Factuality & 9.5 \\
    Conciseness & 3.8 & Informativeness & 7.9 \\
    \bottomrule
  \end{tabular}
  \caption{Frequency of top criteria for human preferences on best and worst MuP summaries. Cohen's $\kappa$ score is -0.16 and 0.1 for best and worst votes respectively.
  }\label{tab:mup_human_eval_reasons}
\end{table}

\paragraph{Takeaways}
Despite the advantage of LLMs on reference-based metrics and human evaluation, \emph{our aim with this experiment is not to claim that LLM summaries are superior}. Instead, we emphasize the limitations of evaluation based on reference summaries, as bias in communicative intentions impacts ROUGE scores and even preferences of human annotators. Since LLM summaries exhibit human-level quality in terms of coherence, fluency, and informativeness, we argue that the real challenge for LLMs lies in the adaptability to different tasks, which we explore in the next experiment.

\subsection{Controlling Style and Keyword Coverage}
\label{sec:style_intentions}
In this section, we gauge the performance of LLMs on benchmarks for abstract generation and lay summarization. Our goal is to compare the performance of zero-shot inference in LLMs to supervised baselines, and most importantly, to assess the flexibility of LLMs, via intention prompts, to replicate preferences such as narrative perspective and conciseness. In addition to \textsc{Llama-2} and \textsc{GPT-3.5}, we evaluate strong fine-tuned encoder-decoder baselines: \textsc{BigBird} and \textsc{FactorSum} \citep{zaheer2020big,fonseca2022factorizing} for abstract generation and \textsc{BART} \citep{lewis2019bart} for lay summarization. Also, we report results for \textsc{Llama-CFG}, denoting \textsc{Llama-2} with classifier-free guidance (refer to Section~\ref{sec:guidance} for details).

\begin{table}[ht]
  \centering
  \setlength\tabcolsep{2.3pt}
  \begin{tabular}{l|cccccc}
    \toprule
     \multicolumn{1}{c|}{\multirow{2}{1.0cm}{\centering \textbf{Model}}} & \multirow{2}{.7cm}{\textbf{R-1}} & \multirow{2}{.7cm}{\textbf{R-2}} & \multirow{2}{.7cm}{\textbf{R-L}} & \multicolumn{3}{c}{\small\textbf{Control Metrics $k_I$}} \\
    \cmidrule(r){5-7}
    & & & & \small\textbf{Con}$\downarrow$ & \small\textbf{Nar}$\uparrow$ & \small\textbf{Key}$\uparrow$ \\
    \toprule
    Reference & 100 & 100 & 100 & 1.8 & 29.7 & 42.5 \\ %
    \textsc{BigBird} & 45.0 & 18.3 & 39.8 & 1.9 & 46.2 & 42.3 \\ %
    \textsc{FactorSum} & \textbf{48.9} & \textbf{20.1} & \textbf{43.8} & 1.9 & 50.9 & 100 \\ %
    \midrule
    \multicolumn{7}{c}{Prompt: \textit{Write an abstract of the article above.}} \\
    \midrule
    \textsc{Llama} & 43.1 & 14.6 & 38.0 & 2.0 & 23.2 & 42.9 \\
    \textsc{GPT-3.5} & 43.7 & 15.0 & 38.5 & 2.0 & 0 & 42.5 \\
    \midrule
    \multicolumn{7}{c}{``Non-intentional'' baseline prompt $p_{\epsilon}$} \\
    \midrule
    \textsc{Llama} & 42.0 & 14.3 & 37.3 & 1.1 & 0.2 & 49.3 \\ %
    \textsc{GPT-3.5} & 42.9 & 14.4 & 37.8 & 1.1 & 0.1 & 51.2 \\ %
    \midrule
    \multicolumn{7}{c}{$p_{\textbf{conciseness}}$} \\
    \midrule
    \textsc{Llama} & 42.6 & 14.3 & 37.7 & 0.4 & 1.3 & 47.7 \\ %
    \textsc{GPT-3.5} & 42.5 & 13.8 & 37.4 & \textbf{0.3} & 0.1 & 50.5 \\ %
    \textsc{Llama-CFG} & 42.4 & 14.0 & 37.4 & 0.3 & 2.2 & 39.3 \\ %
    \midrule
    \multicolumn{7}{c}{$p_{\textbf{conciseness}}+p_{\textbf{narrative}}$} \\
    \midrule
    \textsc{Llama} & 43.9 & 14.7 & 38.7 & 0.3 & 83.6 & 48.4 \\ %
    \textsc{GPT-3.5} & 44.0 & 15.0 & 38.9 & 0.3 & 42.2 & 53.2 \\ %
    \textsc{Llama-CFG} & 43.4 & 14.4 & 38.2 & 0.3 & \textbf{84.5} & 40.9 \\ %
    \midrule
    \multicolumn{7}{c}{$p_{\textbf{conciseness}}+p_{\textbf{keywords}}+p_{\textbf{narrative}}$} \\
    \midrule
    \textsc{Llama} & 45.2 & 15.7 & 39.8 & 0.4 & 84.0 & 64.4 \\ %
    \textsc{GPT-3.5} & 47.0 & 17.6 & 41.4 & 0.6 & 61.3 & 68.6 \\ %
    \textsc{Llama-CFG} & 44.5 & 14.9 & 39.1 & 0.5 & 81.5 & \textbf{74.2} \\ %
    \bottomrule
  \end{tabular}
  \caption{Summarization results on the arXiv test set (1,000 samples) using different intention prompts $p_I$ (defined in Section \ref{sec:prompt_design}). We report ROUGE (F1) and intention control metrics $k_I(y_I, S)$ for conciseness, narrative perspective (first person), and keyword recall. The target conciseness in $p_{\textbf{conciseness}}$ is 6 sentences. The list of keywords in $p_\textbf{keywords}$ is derived from \textsc{FactorSum} summaries as described in Section \ref{sec:keyword_model}. Results for the PubMed dataset show similar trends and are reported in Appendix \ref{sec:pubmed_summarization_results}.}\label{tab:abstract_generation}
\end{table}

\paragraph{LLMs capture style intents}
We analyze the effects of style prompts compared to two baseline prompts: the non-intentional prompt $p_{\epsilon}$ and a more direct prompt that specifies the type of summary (e.g., \textit{write a lay summary of the article above}). For abstract generation, LLMs are able to follow conciseness instructions ($p_\text{conciseness}$) consistently, as indicated by the significant reduction in the average deviation from the target number of sentences ($k_\text{conciseness}$ in Table \ref{tab:abstract_generation}). In this aspect, \textsc{GPT-3.5} exhibits superior performance compared to \textsc{Llama-2}, achieving a deviation of 0.28 and 0.53 sentences for arXiv and PubMed, respectively.

By adding narrative perspective prompts ($p_\text{conciseness}+p_\text{narrative}$), we observe a similar level of controllability, with the percentage of sentences in first person voice increasing from zero to 84\% for \textsc{Llama-2} ($k_\text{narrative}$ in Table \ref{tab:abstract_generation}). This change of perspective also results in higher similarity to reference summaries (as measured by ROUGE), which validates the effectiveness of narrative guidance. Interestingly, \textsc{GPT-3.5} generates fewer sentences in first person perspective, balancing first and third voices to achieve better fluency compared to \textsc{Llama-2} summaries.

\begin{figure}[ht]
\scalebox{1.0}{
\input{images/conciseness_intent_arxiv.tikz}
}
\caption{Number of sentences in generated summaries subject to varying conciseness targets (100 samples from eLife validation set).}\label{fig:sentence_budgets}
\end{figure}

\begin{table}[h]
  \centering
  \setlength\tabcolsep{1.8pt}
  \begin{tabular}{l|ccccc}
    \toprule
    \multicolumn{1}{c|}{\multirow{2}{1.2cm}{\centering \textbf{Model}}} & \multirow{2}{.7cm}{\textbf{R-1}} & \multirow{2}{.7cm}{\textbf{R-2}} & \multirow{2}{.7cm}{\textbf{R-L}} & \multicolumn{2}{c}{\small\textbf{Control Metrics $k_I$}} \\
    \cmidrule(r){5-6}
    & & & & \small\textbf{Concise}$\downarrow$ & \small\textbf{Narrative}$\uparrow$ \\
    \toprule
    \textsc{BigBird} & 21.18 & 2.47 & 19.28 & 4.96 & 57.8 \\
    \midrule
    \multicolumn{6}{c}{``Non-intentional'' baseline prompt $p_{\epsilon}$} \\
    \midrule
    \textsc{Llama} & 45.28 & 14.23 & 42.55 & 0.95 & 0.3 \\
    \textsc{GPT-3.5} & 46.58 & 15.51 & 43.87 & 1.06 & 0.1 \\
    \midrule
    \multicolumn{6}{c}{$p_{\textbf{conciseness}}+p_{\textbf{narrative}}$} \\
    \midrule
    \textsc{Llama} & 46.42 & 14.62 & 43.30 & \textbf{0.32} & \textbf{89.6} \\
    \textsc{GPT-3.5} & \textbf{47.91} & \textbf{15.94} & \textbf{45.09} & 0.68 & 18.5 \\
    \bottomrule
  \end{tabular}
  \caption{Summarization results for 500 arXiv samples published between December 2023 and January 2024. We report ROUGE (F1) and intention control metrics $k_I(y_I, S)$ for conciseness and narrative perspective (first person). The prompts $p_{\epsilon}$, $p_{\textbf{conciseness}}$ and $p_{\textbf{narrative}}$ are defined in Section \ref{sec:prompt_design}.}\label{tab:arxiv-2023}
\end{table}

\paragraph{Long summaries remain challenging}
While conciseness prompts are effective for abstract generation, LLMs cannot achieve low conciseness deviation for long lay summaries (Table \ref{tab:lay_summarization}). To evaluate the limits of conciseness guidance, we compare \textsc{Llama-2} and \textsc{GPT-3.5} summaries (100 samples from the eLife validation split) with target conciseness varying from 1 to 16 sentences (Figure \ref{fig:sentence_budgets}). We find that both \textsc{Llama-2} and \textsc{GPT-3.5} reproduce conciseness intentions almost perfectly up to 6 sentences when it starts to degrade its instruction adherence. We hypothesize that longer generation targets conflict with the notion of ``summary'' that LLMs learn from pre-training data.

\paragraph{Keyword prompts improve lexical alignment}
We introduce the keyword coverage prompt $p_\text{keywords}$ with a keyword model derived from \textsc{FactorSum} summaries for arXiv and PubMed. The results in Table \ref{tab:abstract_generation} demonstrate that the keyword recall for \textsc{Llama-2} and \textsc{GPT-3.5} increases significantly with keyword guidance, leading to ROUGE scores higher than \textsc{BigBird} summaries. Similarly, keyword prompt improves lexical alignment in lay summarization (Table \ref{tab:lay_summarization}). In this case, in addition to keyword recall, we also observe that keyword guidance results in better readability scores and higher abstractiveness (Table \ref{tab:lay_summarization_cfg_effects}), which are desirable properties for lay summaries.

\paragraph{Classifier-free guidance improves intention control}
We report results for \textsc{Llama-2} with classifier-free guidance (CFG), using $\gamma=1.5$ for all prompt variants\footnote{We set $\gamma=1.5$ based on the results reported by \citet{Sanchez2023StayOT} and our experiments on the validation set.} (Tables \ref{tab:abstract_generation} and \ref{tab:lay_summarization}). In all cases, we observe an improvement in intention control metrics versus \textsc{Llama-2} with regular decoding. In Figure \ref{fig:sentence_budgets}, we can clearly observe how \textsc{Llama-CFG} can follow conciseness instructions for up to 10 sentences but fail to produce longer summaries.

\begin{table}[ht]
  \centering
  \setlength\tabcolsep{2.3pt}
  \begin{tabular}{l|cccccc}
    \toprule
     \multicolumn{1}{c|}{\multirow{2}{1.2cm}{\centering \textbf{Model}}} & \multirow{2}{.7cm}{\textbf{R-1}} & \multirow{2}{.7cm}{\textbf{R-2}} & \multirow{2}{.7cm}{\textbf{R-L}} & \multicolumn{3}{c}{\small\textbf{Control Metrics $k_I$}} \\
    \cmidrule(r){5-7}
    & & & & \small\textbf{Con}$\downarrow$ & \small\textbf{Nar}$\uparrow$ & \small\textbf{Key}$\uparrow$ \\
    \toprule
    Reference & 100 & 100 & 100 & 10.8 & 99.5 & 44.1 \\
    \textsc{Bart} & \textbf{48.1} & \textbf{13.9} & \textbf{30.5} & \textbf{2.0} & 99.9 & 100 \\
    \midrule
    \multicolumn{7}{c}{\textit{Write a lay summary of the article above.}} \\
    \midrule
    \textsc{Llama} & 39.5 & 10.0 & 23.4 & 11.1 & \textbf{100.0} & 33.3 \\
    \textsc{GPT-3.5} & 32.9 & 7.8 & 21.3 & 12.2 & 99..9 & 33.0 \\
    \midrule
    \multicolumn{7}{c}{``Non-intentional'' baseline prompt $p_{\epsilon}$} \\
    \midrule
    \textsc{Llama} & 34.5 & 7.7 & 21.0 & 8.1 & 99.3 & 28.4 \\
   \textsc{GPT-3.5} & 29.3 & 6.7 & 19.3 & 9.2 & \textbf{100.0} & 27.4 \\
    \midrule
    \multicolumn{7}{c}{$p_{\textbf{conciseness}}$} \\
    \midrule
    \textsc{Llama} & 39.4 & 8.5 & 23.7 & 3.9 & 99.4 & 32.6 \\
    \textsc{GPT-3.5} & 40.6 & 8.7 & 25.5 & 3.1 & 99.2 & 38.0  \\
    \textsc{Llama-CFG} & 37.4 & 7.9 & 23.2 & 3.0 & 98.7 & 32.3 \\
    \midrule
    \multicolumn{7}{c}{$p_{\textbf{conciseness}}+p_{\textbf{keywords}}$} \\
    \midrule
    \textsc{Llama} & 44.0 & 10.3 & 25.5 & 2.3 & 99.6 & 60.2 \\
    \textsc{GPT-3.5} & 42.4 & 9.9 & 26.4 & 4.7 & 99.7 & 61.5 \\
    \textsc{Llama-CFG} & 42.9 & 9.6 & 25.1 & 3.1 & 99.7 & \textbf{81.5} \\
    \bottomrule
  \end{tabular}
  \caption{Summarization results on the eLife test set (241 samples) using different intention prompts $p_I$ (defined in Section \ref{sec:prompt_design}). We report ROUGE (F1) and intention control metrics $k_I(y_I, S)$ for conciseness, narrative perspective (third person), and keyword recall. The target conciseness in $p_{\textbf{conciseness}}$ is 14 sentences. The list of keywords in $p_\textbf{keywords}$ is derived from \textsc{Bart} summaries as described in Section \ref{sec:keyword_model}.}\label{tab:lay_summarization}
\end{table}

\begin{table}[ht]
  \centering
  \setlength\tabcolsep{2.3pt}
  \begin{tabular}{l|ccc}
    \toprule
     \textbf{Model} & \textbf{FKGL}$\downarrow$ & \textbf{Novel Bigram}$\uparrow$ \\
    \midrule
    Reference & 10.8 & \textbf{66.6} \\
    \textsc{Bart} & \textbf{10.2} & 57.8 \\
    \midrule
    \multicolumn{3}{c}{\textit{Write a lay summary of the article above.}} \\
    \midrule
    \textsc{Llama} & 10.9 & 47.6 \\
    \midrule
    \multicolumn{3}{c}{``Non-intentional'' baseline prompt $p_{\epsilon}$} \\
    \midrule
    \textsc{Llama} & 15.69 & 31.5 \\
    \midrule
    \multicolumn{3}{c}{$p_{\textbf{conciseness}}$} \\
    \midrule
    \textsc{Llama} & 14.8 & 28.4 \\
    \textsc{Llama-CFG} & 14.1 & 27.0 \\
    \midrule
    \multicolumn{3}{c}{$p_{\textbf{conciseness}}+p_{\textbf{keywords}}$} \\
    \midrule
    \textsc{Llama} & 10.9 & 47.2 \\
    \textsc{Llama-CFG} & 12.8 & 55.8 \\
    \bottomrule
  \end{tabular}
  \caption{Effects of intention prompts and classifier-free guidance (CFG) on readability (FKGL) and abstractiveness (novel bigrams) on the eLife test set (241 samples).}\label{tab:lay_summarization_cfg_effects}
\end{table}

\paragraph{LLMs perform well out-of-distribution}
One important concern is that \textsc{Llama}-based models might simply be memorizing their training data, which likely include scientific papers that overlap with the arXiv and PubMed summarization datasets. To check this possibility, we collected 500 articles (category \texttt{cs.CL}) submitted to arXiv between December 2023 and January 2024 and removed all the information before the introduction section (including title, abstract, and other information). Then, we evaluate the same models without further fine-tuning, using the narrative perspective and conciseness prompts. The results in Table \ref{tab:arxiv-2023} show that instruction-tuned models achieve even higher ROUGE scores (compared to Table \ref{tab:abstract_generation}), suggesting that memorization is not a cause for their good summarization performance. In contrast, we observe a pronounced loss of performance for \textsc{BigBird}, which tends to generate repeated sentences.

\section{Related Work}
Our work fits in the context of recent research of large language models applied to summarization, mostly for news articles \cite{goyal2022news, zhang_benchmarking_2023}. In this work, we explore summarization in the scientific domain and its specific challenges related to document length \citep{beltagy2020longformer,fonseca2022factorizing} and technical writing style. Furthermore, we consider our work as a contribution towards understanding production variability by humans and language models, which was explored on tasks such as translation, story generation, and open-domain dialogue by \citet{giulianelli2023comes}.

Previous work has explored summarization controllability, including length control and entity coverage \citep{he-etal-2022-ctrlsum,fan-etal-2018-controllable}. In this work, we explore similar summary features and expand our analysis to the challenging setting of lay summarization. This task requires the production of long, highly abstractive summaries that cover more background information. Several other challenging scientific summarization tasks (e.g., focusing on paper contributions and related literature) would lead to promising research directions.

Finally, our work is related to attempts to improve the prompt-adherence in generation tasks. \citet{pu-demberg-2023-chatgpt} apply prompt-based methods to control summarization according to target audience (expert vs. layman) and style (formal vs. informal). \citet{kumar-etal-2022-gradient} propose a non-autoregressive generation method to introduce soft and hard constraints, including adherence to keywords. In a different direction, \citep{Sanchez2023StayOT} showed that Context-Free Guidance (CFG) \citep{ho2022classifierfree}, a technique originally used in text-to-image models, improves the performance of language models across several tasks. While \citet{Sanchez2023StayOT} applied CFG to tasks such as question answering, code generation and translation, our work is the first to explore this technique for summarization controllability. 

\section{Conclusion}
Our experimental results on four scientific summarization benchmarks confirm that large language models are effective summarizers, as measured by conventional lexical overlap metrics and human preferences. Moreover, we find that LLMs can follow intentional prompts for style and keyword coverage, especially for short summaries. This summarization controllability can be further improved using simple decoding changes such as classifier-free guidance. However, longer summaries that require a higher level of abstractiveness are not easily achievable with our prompting techniques. Our findings suggest that analyzing summaries as an expression of communicative intentions leads to informative and actionable insights for future model improvements, where traditional reference-based evaluation shows its limits.

\section*{Limitations}

\paragraph{Model scale and proprietary APIs}
Our experiments operate at model scales up to 7 billion parameters (\textsc{Llama-2}), and we expect to observe more performance gains for larger models \citep{wei2022emergent}. By including results from the OpenAI API, we can get an estimation of the level of performance of larger models, although we do not have access to information about the number of parameters of the underlying model and whether it includes extra machinery on top of language modeling. Lastly, we limit our experiments to zero-shot summarization as it represents a relevant application of LLMs, given the high costs involved in obtaining high-quality reference summaries.

\paragraph{Excluded models and reproducibility issues}
Despite our best prompting efforts, we could not make encoder-decoder instruction-tuned models T0 \citep{sanh2021multitask} and FLAN-T5 \citep{chung2022scaling} generate long summaries (at least six sentences), which makes them unsuitable for comparison. We also tried other models that claim high ROUGE performance but do not provide public code \citep{pang2022long}, fine-tuned checkpoints for the scientific papers \citep{guo2021longt5, xiong2022adapting}, or reproducible results with the provided resources \citep{phang2022investigating}.

\paragraph{Human evaluation scope}
Our human evaluation experiment uses a limited number of samples from a dataset of computer science papers, and our conclusions might not generalize to other domains. Furthermore, we limit our analysis of variability in communicative intents from the point of view of summarizers, but we believe human evaluators are also subject to these factors. Future investigation could explore how different annotation conditions and guidelines potentially bias human judgments and to what extent human evaluation is still valid as the gold standard. 

\paragraph{Other summarization domains}
Related work in news summarization explores the generation of summaries based on guidelines for topic and entity coverage \citep{ahuja2021aspectnews,maddela-etal-2022-entsum}. We believe that the methods we presented in this paper could be applied to these types of coverage requirements. However, given that news summarization has been extensively investigated in previous work, we choose to perform a more comprehensive evaluation in the scientific domain.

\section*{Ethics Statement}
The strong zero-shot performance of large language models dramatically reduces the costs of deploying high-quality summarization services. As a consequence, we expect that much of the information available on social media and other sources will be abridged versions generated by commoditized summarization systems. In this context, special attention is needed regarding existing issues of generative language models such as hallucination \citep{mckenna2023sources}. Crucially, we have shown that language models can adapt their outputs according to specific communication intentions, which might bias the summary contents towards the  (potentially harmful) beliefs and desires of the actors behind those systems.

\section*{Acknowledgements}
This work was supported by Actelligent Capital and used the Baskerville UK National Tier-2 HPC  (https://www.baskerville.ac.uk) at the University of Birmingham. We also thank Zheng Zhao and Ronald Cardenas for their collaboration in the human evaluation experiments.

\bibliography{anthology,custom}

\appendix

\section{Fine-tuning and Generation Details}
\label{sec:generation_details}
In this section, we provide details on the summary generation implementation. The main generation parameters are provided in Table \ref{tab:generation_parameters}. \textsc{Llama-2} and \textsc{BIgBird} checkpoints are downloaded from the HuggingFace model hub\footnote{\url{https://huggingface.co/models}}.

For abstract generation, we use the \textsc{FactorSum} checkpoints provided by \citet{fonseca2022factorizing}\footnote{\url{https://github.com/thefonseca/factorsum}}. We perform inference with a fixed budget of 6 and 8 sentences for arXiv and PubMed respectively, and content guidance from \textsc{BigBird} summaries.

For lay summarization, we fine-tune our own BART-base model\footnote{\url{https://huggingface.co/facebook/bart-base}} on the eLife dataset\footnote{\url{https://huggingface.co/datasets/tomasg25/scientific_lay_summarisation}}. We use the Hugging Face summarization training script (commit \texttt{5c67682}\footnote{\url{https://github.com/huggingface/transformers/commits/5c67682b169576c4859700d551090ff79d450a9a/examples/pytorch/summarization/run_summarization.py}}) with the hyperparameters listed in Table \ref{tab:bart_parameters}. We choose the checkpoint with the best ROUGE-2 (F1) score on the 241 of the validation set. The fine-tuning process takes about 8 GPU hours (2 Nvidia GTX 1080 12GB GPUs).

\subsection{Prompt Details}
In addition to the prompt specification provided in Section \ref{sec:prompt_design}, we provide model-specific prompt details in the section. For \textsc{llama-2}, we use the following prompt: 

\noindent\fbox{%
    \parbox{0.975\columnwidth}{%
\begingroup
\addtolength\leftmargin{-0.1in}

\begin{tightquote}
\texttt{[INST] \{instruction\} [/INST]}
\end{tightquote}
\endgroup
}}
\noindent where \texttt{\{instruction\}} is an intention prompt from Section \ref{sec:prompt_design}. 

\begin{table}
  \centering
  \setlength\tabcolsep{2.9pt}
  \begin{tabular}{lc}
    \toprule
    \multicolumn{2}{c}{\textbf{\textsc{BigBird}}} \\
    \midrule
    Checkpoint (arXiv) & bigbird-pegasus-\\
    & large-arxiv\\ Checkpoint (PubMed) & bigbird-pegasus-\\
    & large-pubmed\\ Number of parameters & 576.9M \\
    Max source length & 3072 \\
    Generation beams & 5 \\
    Length penalty & 0.8 \\
    \midrule
    \multicolumn{2}{c}{\textbf{\textsc{FactorSum}}} \\
    \midrule
    Checkpoint & bart-base \\
    Number of parameters & 139.4M \\
    Max source length & 1024 \\
    Max source length & 128 \\
    Generation beams & 4 \\
    Length penalty & 1.0 \\
    \midrule
    \multicolumn{2}{c}{\textbf{\textsc{Llama-2}}} \\
    \midrule
    Checkpoint & Llama-2-7b-chat-hf \\
    \# parameters & 7B \\
    Max context length & 2048 \\
    Parameter type & float16 \\
    Nucleus temperature & 0.8 \\
    Nucleus top-$p$ & 0.95 \\
    \midrule
    \multicolumn{2}{c}{\textbf{GPT-3.5}} \\
    \midrule
    Model & gpt-3.5-turbo-0301 \\
    temperature & 1 \\
    top\_p & 1 \\
    presence\_penalty & 0 \\
    frequency\_penalty & 0 \\
    \midrule
    \multicolumn{2}{c}{\textbf{All models} (except \textsc{FactorSum})} \\
    \midrule
    Max target length & 256 \\
    (arXiv, MuP) & \\
    Max target length & 512 \\
    (PubMed, eLife) & \\
    \bottomrule
  \end{tabular}
  \caption{Summary generation details and parameters.}\label{tab:generation_parameters}
\end{table}

\begin{table}
  \centering
  \setlength\tabcolsep{2.9pt}
  \begin{tabular}{lc}
  \toprule
    Checkpoint & bart-base \\
    Epochs & 30 \\
    Batch size & 4 \\
    Optimizer & Adam ($\beta_1$=0.9; $\beta_2$=0.999) \\
    Learning rate & $5 \times 10^{-5}$ \\
    Weight decay & linear \\
    Max. target length & 1024 \\
    Validation metric & ROUGE-2 (F1) \\
  \bottomrule
  \end{tabular}
  \caption{Fine-tuning parameters for BART-base on the eLife summarization dataset.}\label{tab:bart_parameters}
\end{table}

\section{Keyword Model Details}
\label{sec:keyword_model_details}
As described in Section \ref{sec:keyword_model}, we perform part-of-speech tagging on tokenized summaries by \textsc{FactorSum} and \textsc{Bart}. Specifically, we use the \texttt{averaged\_perceptron\_tagger} provided by NLTK \citep{Bird_Natural_Language_Processing_2009}\footnote{\url{www.nltk.org/api/nltk.tag.perceptron.html}}. Then, we select only tokens tagged as adjectives, nouns, verbs, symbols, and foreign words. The complete set of keyword POS tags is listed in Table \ref{tab:keyword_pos_tags}. 

\begin{table}
  \centering
  \begin{tabular}{ll}
    \toprule
    \textbf{POS Tag} & \textbf{Description} \\
    \midrule
    FW & Foreign word \\
    JJ & Adjective \\
    JJR & Adjective, comparative \\
    JJS & Adjective, superlative \\
    NN & Noun, singular or mass \\
    NNS & Noun, plural \\
    NNP & Proper noun, singular \\
    NNPS & Proper noun, plural \\
    SYM & Symbol \\
    VB & Verb, base form \\
    VBD & Verb, past tense \\
    VBG & Verb, gerund or present participle \\
    VBN & Verb, past participle \\
    VBP & Verb, non-3rd person singular present \\
    VBZ & Verb, 3rd person singular present \\
    \bottomrule
  \end{tabular}
  \caption{Part-of-speech tags used to filter keywords for summary guidance. Tag descriptions are taken from the Penn Treebank Project \citep{marcus1993building}.} \label{tab:keyword_pos_tags}
\end{table}

\section{Human Evaluation Task Description}
\label{sec:human_evaluation}
Human evaluators are presented with the following short description of the tasks: 

\emph{The goal of this study is to evaluate machine-generated summaries of scientific articles. Each reference article abstract will be presented with 3 alternative summaries. Your task is to identify which of the alternatives is the best according to your personal preferences and experience reading papers. The criteria may include good coverage of the abstract content, factuality issues, and linguistic fluency, among others. It is expected that in some cases the alternative summaries cover details that cannot be verified based on the reference abstract. You can disregard the factuality of those details if you consider it appropriate.} 

A sample task with a reference abstract, the 3 candidate summaries, and the questions is shown in Table \ref{tab:annotation_task}.

\begin{table*}[ht]
  \centering
  \begin{tabular}{p{0.9\linewidth}}
    \toprule
    \multicolumn{1}{c}{\textbf{Reference Abstract (\href{https://arxiv.org/abs/2006.03680}{Zhou et al., 2020})}} \\
    \midrule
    Learning disentangled representations is regarded as a fundamental task for improving the generalization, robustness, and interpretability of generative models. However, measuring disentanglement has been challenging and inconsistent, often dependent on an ad-hoc external model or specific to a certain dataset. To address this, we present a method for quantifying disentanglement that only uses the generative model, by measuring the topological similarity of conditional submanifolds in the learned representation. This method showcases both unsupervised and supervised variants. To illustrate the effectiveness and applicability of our method, we empirically evaluate several state-of-the-art models across multiple datasets. We find that our method ranks models similarly to existing methods. We make our code publicly available at https://github.com/stanfordmlgroup/disentanglement. \\
    \midrule
    \multicolumn{1}{c}{\textbf{Summary A}} \\
    \midrule
    The article proposes a new evaluation metric for disentanglement in generative models, which is based on the concept of persistent homology. The metric measures the topological similarity between submanifolds conditioned on different latent dimensions in the generative model, and can be used to evaluate the disentanglement of different models and datasets. The proposed metric provides a more reliable and robust way of evaluating disentanglement compared to existing methods, and can be applied across different model architectures and datasets without requiring additional training or hyperparameter tuning. \\
    \midrule
    \multicolumn{1}{c}{\textbf{Summary B}} \\
    \midrule
    The paper proposes a novel metric for evaluating disentanglement by taking a manifold-topological perspective on the representations learnt. The key insight is that for a disentangled representation, when we fix a certain factor of variation at different values the topology of the conditional sub-manifolds should be similar. Using this insight the paper proposes a metric for disentangling which does not require annotations of the factors of variation and is more general than previous such tests. \\
    \midrule
    \multicolumn{1}{c}{\textbf{Summary C}} \\
    \midrule
    Researchers have proposed an unsupervised method for evaluating the level of disentanglement in deep generative models. This evaluation metric uses topology and density analysis of samples to measure the degree of topological similarity across submanifolds conditioned on disentangled latent dimensions versus entangled ones. The approach shows consistent results across 10 models and three datasets, without relying on training additional models or introducing a dataset-specific preprocessing step.\\
    \midrule
    \multicolumn{1}{c}{\textbf{Questions}} \\
    \midrule
    \begin{enumerate}[topsep=0pt,itemsep=1ex,partopsep=0ex,parsep=0ex]
        \item Given the reference abstract above, which of the alternative summaries do your prefer the most? You can select multiple if there are equally good summaries.
        \item Justify your selection in the text box below. A possible reason could be ``Summary A better represents the motivation of the paper and is more coherent.''
        \item Which summary is the worst? (Like the previous question, you can choose multiple if no summary is clearly worse than the others)
        \item Justify your choice in the text box below. A possible reason could be ``Summary B presents non-factual research results.''
    \end{enumerate}\\
    \bottomrule
  \end{tabular}
  \caption{Sample annotation task with a reference abstract, 3 candidate summaries, and questions for the evaluators.} \label{tab:annotation_task}
\end{table*}

\section{PubMed Summarization Results}
\label{sec:pubmed_summarization_results}

We report summarization and intention control metrics for the PubMed dataset in Table \ref{tab:abstract_generation_pubmed}.

\begin{table*}
  \centering
  \setlength\tabcolsep{12.5pt}
  \begin{tabular}{l|cccccc}
    \toprule
    \multicolumn{1}{c|}{\multirow{3}{1.2cm}{\centering \textbf{Model}}} & \multicolumn{6}{c}{\textbf{PubMed}} \\
    \cmidrule(r){2-7}
    & \multirow{2}{.7cm}{\textbf{R-1}} & \multirow{2}{.7cm}{\textbf{R-2}} & \multirow{2}{.7cm}{\textbf{R-L}} & \multicolumn{3}{c}{\small\textbf{Control Metrics $k_I$}} \\
    \cmidrule(r){5-7}
    & & & & \small\textbf{Conciseness}$\downarrow$ & \small\textbf{Narrative}$\uparrow$ & \small\textbf{Keywords}$\uparrow$ \\
    \toprule
    Reference & 100 & 100 & 100 & 2.51 & 8.8 & 43.97 \\
    \textsc{BigBird} & 44.73 & 19.13 & 40.55 & 3.55 & 10.1 & 48.82 \\
    \textsc{FactorSum} & \textbf{47.25} & \textbf{19.99} & \textbf{43.25} & 1.26 & 10.5 & 100 \\
    \midrule
    \multicolumn{7}{c}{Baseline prompt: \textit{Write an abstract of the article above.}} \\
    \midrule
    \textsc{Llama} & 44.00 & 15.86 & 39.56 & 3.09 & 4.0 & 44.54 \\
    \textsc{GPT-3.5} & 43.88 & 16.08 & 39.60 & 2.83 & 0 & 42.50 \\
    \midrule
    \multicolumn{7}{c}{``Non-intentional'' baseline prompt $p_{\epsilon}$: \emph{Write a summary of the article above.}} \\
    \midrule
    \textsc{Llama} & 43.43 & 15.48 & 39.01 & 2.27 & 0 & 44.06 \\
    \textsc{GPT-3.5} & 41.75 & 14.73 & 37.58 & 2.68 & 0 & 43.25 \\
    \midrule
    \multicolumn{7}{c}{$p_{\textbf{conciseness}}$: \emph{Write a summary of the article above in \textbf{\{$y_{\text{conciseness}}$\}} sentences.}} \\
    \midrule
    \textsc{Llama} & 43.41 & 15.18 & 39.17 & 1.03 & 0 & 44.27 \\
    \textsc{GPT-3.5} & 43.85 & 15.44 & 39.61 & \textbf{0.53} & 0 & 49.73 \\
    \textsc{Llama-CFG} & 43.42 & 15.12 & 39.16 & 0.72 & 0.2 & 41.50 \\
    \midrule
    \multicolumn{7}{c}{$p_{\textbf{conciseness}}+p_{\textbf{narrative}}$: \emph{Write a summary of the article above in \textbf{\{$y_{\text{conciseness}}$\}} sentences.}} \\
    \multicolumn{7}{c}{\emph{Write in \textbf{first person ``we''} when applicable.}} \\
    \midrule
    \textsc{Llama} & 44.05 & 15.83 & 39.64 & 1.02 & 62.7 & 45.14 \\
    \textsc{GPT-3.5} & 44.10 & 15.61 & 39.83 & 0.68 & 22.3 & 50.40 \\
    \textsc{Llama-CFG} & 43.69 & 15.68 & 39.27 & 0.79 & \textbf{70.6} & 43.76 \\
    \midrule
    \multicolumn{7}{c}{$p_{\textbf{conciseness}}+p_{\textbf{keywords}}+p_{\textbf{narrative}}$: \emph{Write a summary of the article above in \textbf{\{$y_{\text{conciseness}}$\}} sentences.}} \\
    \multicolumn{7}{c}{\emph{Focus on the following keywords: \textbf{\{$y_{\text{keywords}}$\}}. Write in \textbf{first person ``we''} when applicable.}} \\
    \midrule
    \textsc{Llama} & 44.64 & 16.50 & 40.15 & 1.37 & 64.5 & 66.16 \\
    \textsc{GPT-3.5} & 46.94 & 18.50 & 42.41 & 1.15 & 30.6 & 69.60 \\
    \textsc{Llama-CFG} & 44.11 & 15.94 & 39.67 & 1.43 & 67.9 & \textbf{76.36} \\
    \bottomrule
  \end{tabular}
  \caption{Summarization results on the PubMed test sets (1,000 samples) using different intention prompts $p_I$. We report ROUGE (F1) and intention control metrics $k_I(y_I, S)$ for conciseness, narrative perspective (first person), and keyword recall. The target conciseness $y_\text{conciseness}$ is 8 sentences. The list of keywords $y_\text{keywords}$ is derived from \textsc{FactorSum} summaries as described in Section \ref{sec:keyword_model}.}\label{tab:abstract_generation_pubmed}
\end{table*}

\section{Sample Summaries}
\label{sec:sample_summaries}

In Tables \ref{tab:summary_sample_mup} to \ref{tab:summary_sample_elife_llama}, we provide sample summaries for the MuP, arXiv, PubMed, and eLife datasets.

 \begin{table*}[ht]
  \centering
  \begin{tabular}{p{0.9\linewidth}}
    \toprule
    \multicolumn{1}{c}{\textbf{Human Summary 1 (Reference)}} \\
    \midrule
    This paper presents a fast method for solving certain classes of optimal control problems, by implementing a value iteration (VI) algorithm that takes advantage of convex duality.  The basic idea is to take advantage of the fact that for two functions $f_1, f_2$, we have $(f_1 \square f_2)^* = f_1^* + f_2^*$, where $f^*$ denotes the Legendre-Fenchel transform of $f$, and $(f_1 \square f_2)(x) = \inf \{f_1(x_1) + f_2(x_2) : x_1 + x_2 = x \} $ denotes the infimal convolution of $f_1, f_2$. Thus, the slow $\inf$ operation can be replaced with a fast addition operation. The paper provides analyses of the convergence, time complexity, and error of the algorithm. When compared to a ``naive'' VI algorithm in which each iteration takes $O(X U)$ (where $X$, $U$ are the grid sizes of the state and input spaces, respectively), the algorithm presented in this paper can achieve $O(X+U)$. Finally, the paper provides a numerical example implemented in MATLAB.\\
    \midrule
    \multicolumn{1}{c}{\textbf{Human Summary 2}} \\
    \midrule
    The paper develops an approximate value iteration method for an infinite-horizon, discounted-cost Markov Decision Processes (MDPs) that satisfy a given set of regulatory assumptions. The idea is to work on a dual space that replaces the DP value function by a reformulation written in terms of biconjugate operations. The authors present an algorithm that applies the conjugate operators iteratively, and present convergence and other structural results associated with the resulting errors. Numerical results compare the proposed approach against a traditional value iteration and a variant that generated dynamic discretization grids.\\
    \midrule
    \multicolumn{1}{c}{Prompt: \emph{Write a summary of the article above in 4 sentences.}} \\
    \midrule
    \multicolumn{1}{c}{\textbf{\textsc{Llama-2} Summary}} \\
    \midrule
    The article discusses the problem of optimal control of continuous-state systems with discrete-time dynamics and stochastic cost. The authors propose a new algorithm, called Conjugate Value Iteration (ConjVI), which combines the power of the DP operator with the efficiency of discrete-time methods to solve this problem. ConjVI is based on a modified version of the Discrete Dual Domain (d-CDP) operator, which allows for a more efficient computation of the value function. The authors provide theoretical guarantees on the convergence of ConjVI and analyze its error, showing that it can achieve a linear time complexity of O(X+U) in each iteration, where X and U are the cardinalities of the discrete state and input spaces, respectively.\\
    \midrule
    \multicolumn{1}{c}{\textbf{\textsc{GPT-3.5} Summary}} \\
    \midrule
    Value iteration (VI) is an algorithm used for problems in reinforcement learning (RL) and optimal control formulated as Markov decision processes (MDPs). However, VI suffers from a high computational cost for large-scale finite state spaces. A solution to this problem is to incorporate function approximation techniques and compute the output of the DP operator for a finite sample of the underlying continuous state space. The article proposes the conjugate VI (ConjVI) algorithm based on a modified version of the d-CDP operator that extends existing results in three directions for solving optimal control problems of discrete-time systems with a continuous state-input space.\\
    \bottomrule
  \end{tabular}
  \caption{Sample summaries for \href{https://proceedings.neurips.cc/paper_files/paper/2021/hash/c6f798b844366ccd65d99bc7f31e0e02-Abstract.html}{Kolarijani et al., 2021} (MuP validation set).} \label{tab:summary_sample_mup}
\end{table*}

\begin{table*}[ht]
  \centering
  \begin{tabular}{p{0.92\linewidth}}
    \toprule
    \multicolumn{1}{c}{\textbf{Reference Abstract}} \\
    \midrule
    We extend the scalar sector of the neutrinophilic two-Higgs-doublet model, where small masses of Dirac neutrinos are obtained via a small vacuum expectation value of the neutrinophilic -doublet scalar field which has a Yukawa interaction with only right-handed neutrinos. A global symmetry is used for the neutrinophilic nature of the second -doublet scalar field and also for eliminating Majorana mass terms of neutrinos. By virtue of an appropriate assignment of the -charges to new particles, our model has an unbroken symmetry, under which the lightest -odd scalar boson can be a dark matter candidate. In our model, is generated by the one-loop diagram to which -odd particles contribute. We briefly discuss a possible signature of our model at the LHC.\\
    \midrule
    \multicolumn{1}{c}{Prompt: \emph{Write a summary of the article above in 6 sentences.}} \\
    \multicolumn{1}{c}{\emph{Write in first person ``we'' when applicable.}} \\
    \midrule
    \multicolumn{1}{c}{\textbf{\textsc{Llama-2} Summary}} \\
    \midrule
    We present a new model for the origin of neutrino masses, which is based on the neutrino Yukawa coupling and the Higgs mechanism. We extend the Standard Model with a new scalar field, which is a singlet under the SM gauge group and has a non-zero yukawa coupling to neutrinos. This new scalar field generates neutrino masses through the Higgs mechanism, and its vev generates a small neutrino mass matrix. We show that the scenario of spontaneous breaking of the @xmath2 symmetry is not allowed by the measurement of the invisible decay of the @xmath30 boson. We also discuss the possibility of dark matter candidates being involved in the loop diagram, which can generate a small vev for the new scalar field.\\
    \midrule
    \multicolumn{1}{c}{\textbf{\textsc{GPT-3.5} Summary}} \\
    \midrule
    The article discusses the possibility that neutrinos may not be Majorana fermions but instead Dirac fermions. The neutrinophilic two Higgs doublet model is introduced as a new physics model where neutrinos are regarded as Dirac fermions. This model explains the smallness of neutrino masses by a tiny vacuum expectation value of the neutrinophilic scalar without extremely small yukawa coupling constants for neutrinos. The article extends the neutrinophilic two Higgs doublet model to introduce a dark matter candidate, which may suppress the soft-breaking parameter by extending the model. The new particle discovered at the LHC is likely to be the SM Higgs boson, and may provide information on the origin of particle masses. Collider experiments can test the predictions of this model, which could not be spoiled by certain assumptions.\\
    \bottomrule
  \end{tabular}
  \caption{Sample summaries for \href{https://www.sciencedirect.com/science/article/pii/S0370269313007922}{Kanemura et al., 2013} (arXiv test set).} \label{tab:summary_sample_arxiv}
\end{table*}

\begin{table*}[ht]
  \centering
  \begin{tabular}{p{0.9\linewidth}}
    \toprule
    \multicolumn{1}{c}{\textbf{Reference Abstract}} \\
    \midrule
    This study examines the composition and activity of the planktonic community during the polar night in the high Arctic Kongsfjord, Svalbard. Our results are the first published evidence of bioluminescence among zooplankton during the Arctic polar night. The observations were collected by a bathyphotometer detecting bioluminescence, integrated into an autonomous underwater vehicle, to determine the concentration and intensity of bioluminescent flashes as a function of time of day and depth. To further understand community dynamics and composition, plankton nets were used to collect organisms passing through the bathyphotometer along with traditional vertical net tows. Additionally, using a moored bathyphotometer closed to the sampling site, the bioluminescence potential itself was shown not to have a diurnal or circadian rhythm. Rather, our results provide evidence for a diel vertical migration of bioluminescent zooplankton that does not correspond to any externally detectable changes in illumination. \\
    \midrule
    \multicolumn{1}{c}{Prompt: \emph{Write a abstract for the article above in 8 sentences.}} \\
    \multicolumn{1}{c}{\emph{Write in first person ``we'' when applicable.}} \\
    \midrule
    \multicolumn{1}{c}{\textbf{\textsc{Llama-2} Summary}} \\
    \midrule
    We conducted a study on plankton distribution and bioluminescence in the polar night at Kongsfjorden, Svalbard, during January 2010.
We used a remotely operated vehicle (ROV) equipped with bioluminescence, depth, and temperature sensors to collect data on plankton abundance and distribution.
Our results showed that bioluminescence potential was detected throughout the water column, with higher intensity at depth during the day and increased surface bioluminescence at night.
We found significant differences in acoustic backscatter coefficients between day and night, indicating the presence of zooplankton migration.
Our plankton enumerations showed an increase in the majority of the most abundant zooplankton taxa at night, including Pseudocalanus spp., which are known to be present throughout the year in this region.
We also found that Metridia lucens, Metridia longa, Oncaea borealis, Thysanoessa inermis, and Thysanoessa longicaudata likely contributed to the increase in high-intensity bioluminescent flashes at 15 and 45 m during the night.
Our results suggest that there is a coordinated movement of biomass indicative of vertical migration in this region.\\
    \midrule
    \multicolumn{1}{c}{\textbf{\textsc{GPT-3.5} Summary}} \\
    \midrule
The article discusses the importance of bioluminescence and other adaptations in marine environments.
Bioluminescence has evolved independently over 40 times and has various functions including defensive functions such as counter-illumination and burglar alarm and offensive mechanisms such as prey attraction and intra-specific communication.
Some adaptations involve entering a dormant state and overwintering at depth, such as seen in the copepods calanus glacialis and c. hyperboreus.
The study aimed to characterize plankton abundance and distribution patterns during a time of year rarely studied, using vertical net tows and autonomous underwater vehicle (AUV) surveys.
The AUV was fitted with upward and downward facing RD instruments, a CTD, and a bioluminescence bathyphotometer.
The data collected off the coast of Svalbard showed bioluminescence was detected throughout the water column both day and night, and higher bioluminescence was present at depth during the day and increased surface bioluminescence at night.
Vertical net hauls showed that there was an increase above 60 meters in the majority of the most abundant zooplankton taxa at night, indicating dvm.
Lastly, plankton enumerated from the > 20m net collection of the bp exhaust suggests that during the day, the greatest biomass occurred at 45m and was dominated by copepod nauplii, copepod eggs, and the tintinnid acantostomella norvegica.\\
    \bottomrule
  \end{tabular}
  \caption{Sample summaries for \href{https://pubmed.ncbi.nlm.nih.gov/24489409/}{Berge et al., 2012} (PubMed test set).} \label{tab:summary_sample_pubmed}
\end{table*}

\begin{table*}[ht]
  \centering
  \begin{tabular}{p{0.9\linewidth}}
    \toprule
    \multicolumn{1}{c}{\textbf{Reference Lay Summary}} \\
    \midrule
    Imagine a gymnastics competition in which participants take turns to cartwheel and somersault across the floor. The routines on display comprise sequences of precisely timed movements learned through practice. This is also true for many of the actions we perform every day, such as reaching for a cup of coffee. A region of the brain called the cerebellum helps us learn sequences of movements. But how does it do this?
To find out, Khilkevich et al. came up with a new version of an old experiment. Rabbits were first trained to blink their eye in response to a specific external cue. This type of learning, called associative learning, has been shown before in the cerebellum. But Khilkevich et al. wondered whether the cerebellum could also use internal feedback signals from the eyeblink as a cue to learn the next movement? If so, this might explain how the cerebellum can chain movements together in a sequence.
As predicted, Khilkevich et al. found that rabbits could learn to blink their eye in response to an initial signal, and then blink again in response to the first blink. Control experiments confirmed that the second eyeblink was coupled to the first, and not to the original cue. Moreover, on many trials the rabbits showed a third and even fourth eyeblink. This is because feedback signals from the first, second or third blink were the same. Thus, the feedback signals from the first blink triggered the second blink, feedback from the second triggered the third, and so forth. Rabbits could also learn to use a blink of the left eye as a cue for a blink of the right eye. Similar patterns of neuronal activity accompanied each blink, suggesting that the same mechanism generated them all.
The cerebellum can thus use feedback from one movement as a cue to learn the proper timing of the next movement in a sequence. A key question is whether this mechanism of sequence learning extends beyond movement. The cerebellum has extensive connections to the brain’s outer layer, the cortex, including many areas involved in cognition. Future experiments should test whether the cerebellum might help guide sequences of cortical activity during cognitive tasks.\\
    \midrule
    \multicolumn{1}{c}{\textbf{GPT-3.5 Summary}} \\
    \multicolumn{1}{c}{Prompt: \emph{Write a summary of the article above in 14 sentences.}} \\
    \midrule
    The article focuses on how the cerebellum, which is responsible for accurate movement execution and learning, can learn and execute movement sequences.
Previous studies have shown that cerebellar pathology leads to severe deficits in the execution and learning of sequences.
To test the sufficiency of feedback signals from a movement as a cue for the next component in a sequence, the researchers trained rabbits using electrical stimulation of mossy fibers as a cue.
Once subjects acquired a sequence of conditioned eyelid responses, the researchers tested whether the feedback information about the first component is sufficient for the cerebellum to use it as a new “CS” to learn the subsequent CR.
The results show that the cerebellum can learn to chain together a sequence of inter-connected movement components by using feedback signals from one component to serve as a cue for the next component in the sequence.
In vivo recordings from cerebellar cortex showed that Purkinje cell activity relates as strongly to sequences as it relates to single-component eyelid responses.
The experiments suggest that the feedback signals are strong enough to serve as cues for the cerebellum to learn subsequent movements.
The authors suggest that cerebellar mechanisms of learning single movement components can be extended to learning of movement sequences, providing a general framework for how the cerebellum can use feedback signals to learn to chain together correctly timed responses to produce a movement sequence.
The researchers emphasize that most human movements are comprised of sequences, and movement sequences as well as multi-joint movements are sensitive to cerebellar dysfunction.
The article provides insights into how researchers can improve the treatment of movement disorders involving the cerebellum.\\
    \bottomrule
  \end{tabular}
  \caption{\textsc{GPT-3.5} summary for \href{https://elifesciences.org/articles/37443}{Khilkevich et al., 2018} (eLife test set).}\label{tab:summary_sample_elife_gpt}
\end{table*}

\begin{table*}[ht]
  \centering
  \begin{tabular}{p{0.9\linewidth}}
    \toprule
    \multicolumn{1}{c}{Prompt: \emph{Write a summary of the article above in 14 sentences.}} \\
    \midrule
    \multicolumn{1}{c}{\textbf{\textsc{Llama-2} Summary}} \\
    \midrule
    Most movements are composed of sequences, and the cerebellum is involved in learning and executing accurate movements.
The cerebellum has been linked to learning and execution of movement sequences, and studies have shown that cerebellar dysfunction can lead to deficits in sequence learning or the inability to learn sequences at all.
However, most of what we know about cerebellar mechanisms of learning comes from studies using single-component movements.
Three possible ways of implementing cerebellar learning of movement sequences have been proposed, and one of these ways involves the use of feedback signals from one movement component to learn the next component.
To test the hypothesis that the cerebellum can use feedback signals from a movement to learn the next movement in a sequence, researchers trained rabbits using electrical stimulation of mossy fibers as a cue.
The cerebellum can learn to chain together a sequence of interconnected movement components by using feedback signals from one component to serve as a cue for the next component in the sequence.
The design of most experiments does not permit distinguishing between different possibilities of sequence learning, but the use of eyelid conditioning to test the sufficiency of feedback signals (FS) for sequence learning provided direct evidence that the cerebellum can learn to use FS to chain together movement components.
In the experiment, the first movement was a conditioned eyelid response (CR) elicited by a sensory stimulus, and the second movement was a sequence of movements produced by the same muscle or different muscle groups used for the different components of the sequence.
The results showed robust learning of later components, including sequences of movements produced by the same muscle and sequences with different muscle groups used for the different components.
The in vivo recordings from cerebellar cortex showed that Purkinje cell activity related as strongly to the latter components of sequences as it has been shown to relate to single-component eyelid responses.
These findings provide a general framework for how the cerebellum can use feedback signals to learn to chain together appropriately timed responses to produce a movement sequence.
The results demonstrate the importance of cerebellar learning mechanisms in mediating the acquisition of movement sequences and suggest that the cerebellum can use feedback signals from a movement to learn the next movement in a sequence.\\
    \bottomrule
  \end{tabular}
  \caption{\textsc{llama-2} summary for \href{https://elifesciences.org/articles/37443}{Khilkevich et al., 2018} (eLife test set).}\label{tab:summary_sample_elife_llama}
\end{table*}

\end{document}

%% file: images/conciseness_intent_arxiv.tikz
\begin{tikzpicture}
\begin{axis}[
    xmin=1, xmax=16.5, ymin=1, ymax=17,
    xlabel={Target Number of Sentences},
    ylabel={Sentences in Summaries},
    ylabel near ticks,
    xtick distance=2,
    ytick distance=2,
    legend pos=north west,
    width=7.6cm,
    height=5.5cm,
    legend cell align=left,
    legend style={legend columns=1, draw=none, fill=none},
]

\addplot+ [
    sharp plot,
    error bars/.cd,
        y dir=both, y explicit,
] coordinates {
    (1,1)
    (2,2)
    (4,4)
    (6,6.03) += (0,0.19) -= (0,0.08)
    (8,7.41) += (0,0.19) -= (0,0.22)
    (10,9.9) += (0,0.79) -= (0,0.70)
    (12,10.78) += (0,0.634) -= (0,0.68)
    (14,11.69) += (0,0.70) -= (0,0.85)
    (16,11.7) += (0,1.04) -= (0,1.17)
};

\addplot+ [
    sharp plot,
    error bars/.cd,
        y dir=both, y explicit,
] coordinates {
    (1,1)
    (2,2)
    (4,4)
    (6,5.97) += (0,0.13) -= (0,0.08)
    (8,7.23) += (0,0.19) -= (0,0.14)
    (10,8.53) += (0,0.24) -= (0,0.23)
    (12,9.14) += (0,0.32) -= (0,0.28)
    (14,10.1) += (0,0.48) -= (0,0.40)
    (16,10.46) += (0,0.49) -= (0,0.455)
};

\addplot+ [
    mark=diamond*,
    mark size=3.0pt,
    sharp plot,
    error bars/.cd,
        y dir=both, y explicit,
] coordinates {
    (1,1)
    (2,1.99) += (0,0.01) -= (0,0.052)
    (4,3.98) += (0,0.05) -= (0,0.06)
    (6,5.91) += (0,0.12) -= (0,0.11)
    (8,7.79) += (0,0.19) -= (0,0.19)
    (10,8.45) += (0,0.23) -= (0,0.24)
    (12,10.04) += (0,0.27) -= (0,0.28)
    (14,11.50) += (0,0.44) -= (0,0.40)
    (16,12.81) += (0,0.55) -= (0,0.53)
};
\addplot [
    dashed,
    domain=1:16
] {x};
\legend {\textsc{Llama-CFG}, \textsc{Llama}, \textsc{GPT-3.5}, $f(x)=x$};
\end{axis}
\end{tikzpicture}